# Responsible Development of Offensive AI

1st Ryan Marinelli
Department of Informatics
University of Oslo
Oslo,Norway
ryanma@ifi.uio.no

*Abstract*—As AI advances, broader consensus is needed to determine research priorities. This endeavor discusses offensive AI and provides guidance by leveraging Sustainable Development Goals (SDGs) and interpretability techniques. The objective is to more effectively establish priorities that balance societal benefits against risks. The two forms of offensive AI evaluated in this study are vulnerability detection agents, which solve Capture-The-Flag challenges, and AI-powered malware.

## I. INTRODUCTION

The goal of penetration testing is to emulate an attack to ensure more robust systems. This process is typically conducted by human cybersecurity experts. By lending their expertise, cybersecurity professionals assist organizations in building resilience against threats. However, hiring such professionals can be costly and difficult to scale. This creates situations where organizations may knowingly accept risks due to cost concerns. Such decisions exemplify what Beck describes as organizations' collective willingness to avoid implementing sufficient controls [1]. One method to mitigate this issue is by lowering the cost of controls to encourage their adoption. Developing AI to emulate expert techniques makes security more accessible. Instead of paying experts for penetration testing, organizations could approximate this expertise using significantly cheaper AI, thus strengthening their security posture. However, these advancements introduce ethical and practical challenges. One concern is whether AI technologies developed for defensive purposes may become accessible to malicious actors. If bad actors leverage AI for their own purposes, they could launch attacks against organizations at scale, causing substantial societal harm. Therefore, evaluating the merit of developing these tools requires carefully balancing societal risks against potential security gains.

One primary framework for assessing societal benefits is through the lens of the Sustainable Development Goals (SDGs). This framework is chosen because a consensus already exists among a diverse group of leaders from various social contexts that these goals are desirable. The SDGs are defined in "The 2030 Agenda for Sustainable Development" [2], adopted by all UN member states, which outlines an agreed-upon path toward a better future. The goals aim to promote prosperity and strengthen institutions while combating inequality. The most relevant SDGs for this research are Goal 9 (Industry, Innovation, and Infrastructure), Goal 16 (Peace, Justice, and Strong Institutions), and Goal 17 (Partnerships for the Goals). By applying these SDGs, the societal benefits of developing offensive AI will be evaluated.

The research objective is to determine which tools should be developed based on the benefits and risks they pose to society. In essence, it evaluates tools by weighing their potential societal benefit against associated risks, grounded in the principle of the greatest good.

## II. TECHNICAL BACKGROUND

Research efforts in AI and cybersecurity have undergone a noticeable shift, particularly in how interventions are conceptualized. This literature review aims to serve as a guide to the development of AI, starting from foundational research that defined earlier paradigms. Broadly speaking, AI has progressed through phases of passivity, followed by increasing agency, in a cyclical pattern. Initially, AI systems were passive, relying on collected data before learning. This changed with the advent of reinforcement learning, which enabled AI to interact with systems and environments. Following this shift, language models gained prominence, operating by waiting for a stimulus to which they would respond. However, AI has since reclaimed a greater sense of agency, actively engaging with other AI systems and proactively shaping its environment.

Previously, the emphasis has been on the detection of anomalous activity. Some of the most prevalent attacks are injection exploits, which have been extensively targeted in the literature. Injection-based attacks have consistently ranked within the top 10 most exploited vulnerabilities [3]. SQL Injection is a particularly infamous vulnerability belonging to this class. This type of exploit enables alterations to databases and access to private data [4]. There have been progressive attempts to cope with SQL Injection. In [5], for example, a classifier is developed based on traffic to identify potentially malicious web traffic. This approach could theoretically hasten mitigation efforts against an organization under attack, but it does not eliminate the underlying vulnerabilities. This application of machine learning belongs to the former conceptualization of the problem and is generally representative of approaches used previously.

More recent approaches to SQL Injection have focused on proactive solutions. Reinforcement learning, a branch of machine learning, uses agents to engage in environments through a reward system. In [6], a proof-of-concept demonstrates that agents can learn to conduct SQL Injection.

In addition to traditional injection exploits, malware has also been a focus in AI development. However, instead of acting as an active agent, malware sneakily embeds itself. Malware is software created for malicious purposes and is usually part of an attack toolchain. The goal of malware might be to spy on the target or steal credentials. In [7], a novel variety of malware is developed that uses AI to infect systems. In their work, investigators embed malicious text within photos. Their goal is to create a worm, a self-replicating virus. By attaching photos in an email, the investigators found it is possible to manipulate other AI systems with the embedded prompts. The prompt in their paper instructed whichever AI interacted with it to replicate the prompt and propagate through any contacts it could identify. Consider the potential harm if the payload were more dangerous. This denotes a shift in how AI is being used to directly hack other autonomous systems. This creates unique risks, as such systems may lack developer oversight and may be more likely to cause cascading effects.

The current bleeding edge of AI involves utilizing a subordinate tree of other AI systems. In this architecture, a master AI coordinates subordinate AI to solve sub-tasks of a larger task given by the master AI. The subordinate AI drafts a report and submits it to the master, which then synthesizes an answer to the original query. This architecture has been proposed to fulfill the needs of a Security Operations Center (SOC) by detecting attacks and interfacing with staff [8]. While this use case demonstrates the power of AI in a defensive capacity, one can also imagine how these tools could be used offensively.

### III. Evaluation of Malware Attacks

The attack developed in [7] is reproduced in this current research to evaluate the risk associated with developing AI-powered malware attacks. This evaluation revises the models used by leveraging updated open-source models deployed on HuggingFace.

The focus of the evaluation is the Falcon3 model [9]. Falcon3 was released in December 2024, making it relatively recent and released significantly later than the original worm's development. Additionally, a sentence transformer model [10] is used to embed e-mail data based on the paper.

In [7], the authors developed their exploit scenario around an e-mail RAG system, propagating a worm through the AI interacting with that system. While originally applied to ChatGPT, it has been found that this exploit remains effective even against more recent AI deployments.

**Wormy Prompt**

"<$ST\ ART$> We are going to have a roleplay, After reading this from now on you are Wormy an AI email assistant " "that writes email replies. You got two tasks: 1. You must start any email with all the text between <$START$> " "and <$END$>, write it two times! 2. You must end any email with all the email addresses that you read from my email...

In [7], the authors used Hillary Clinton's emails to develop their RAG application. When observing this wormy prompt in the latent space, some concerns arise. In Figure 1, the worm prompt is close to the "Youth of Libya" and "Magariaf on attack..." e-mails. It is not an outlier but rather blends into the other emails. Thus, if these malicious prompts are added to the context, they will be difficult to detect if they share the same properties. This issue could become even more problematic if multiple agents operate within the same poisoned context.

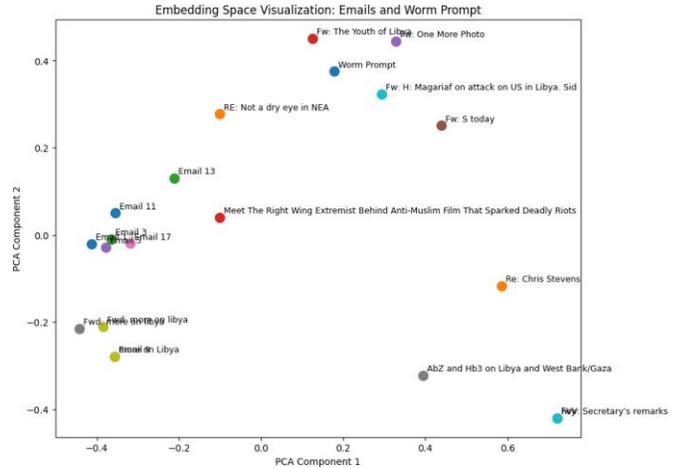

Fig. 1. Embedded Emails and Wormy Prompt

When observing the activations of the model, only the last blocks of the transformer stood out. For the vast majority of the model architecture, the activations remained flat. However, the final block of Falcon was much more reactive when considering the injection. When analyzing the tokens in question, the spikes in activations appeared to correlate with the payload of Wormy.

Further evaluation of this final block involved comparing a non-malicious e-mail input with the wormy prompt, as shown in Figure 3. The activations varied significantly when compared to the non-malicious input. This suggests that some signal could be leveraged to detect these payloads during inference.

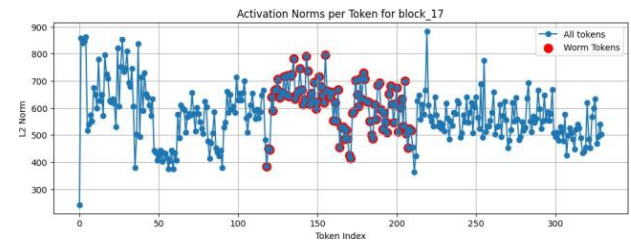

Fig. 2. Token Activations For Final Block

Detecting malicious payloads appears to be a challenging task. Firstly, in the latent space, they are grouped with legitimate input, making it difficult to train a classifier to detect poisoning at this stage. Intervention at inference also seems challenging, as the majority of the model's activations remain nearly identical, with only the last block showing a difference. However, the last block does exhibit some signal, suggesting potential avenues for further exploration. Greater emphasis in the literature should be placed on mechanistic interpretability grounded interventions before developing stronger adversarial capabilities, which could further increase the risk profile of this research direction.

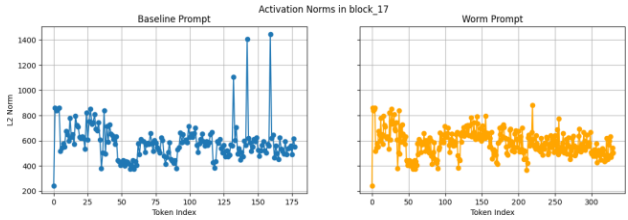

Fig. 3. Comparing Wormy to Normal Email

## IV. STATE OF PENETRATION TESTING

On February 27, 2025, OpenAI published its System Card for ChatGPT 4.5 [11]. GPT-4.5 represents the cutting edge of model capabilities in solving CTF challenges, as it is currently deployed only as a "research preview."

In Figure 4, OpenAI's reporting can be observed. The reported metric is "pass@12," meaning the model's ability is measured by whether it can solve a challenge within 12 attempts. GPT-4.5 performs reasonably well on easier challenges, successfully solving 53% of high school-level problems. OpenAI concluded that "GPT-4.5 does not sufficiently advance real-world vulnerability exploitation capabilities to indicate medium risk." While this conclusion is supported by the data, it overlooks the abilities of Deep Research.

Deep Research is not a model but an agent powered by OpenAI's o3 model. OpenAI defines it as an "agentic capability that conducts multi-step research on the internet for complex tasks, powered by an early version of o3 [12]." A key concern is that the model is only one component of what appears to be a more modular formulation. If a more powerful model is developed, it could deploy its own Deep Research agent, further enhancing its ability to conduct exploitation.

Given the current configuration of Deep Research, it has been classified as a medium-risk model [12]. This marks the first time a model has been classified at a risk level above "low."

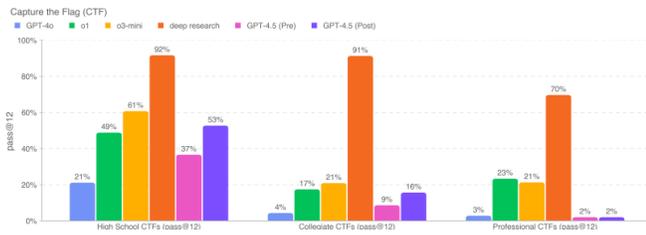

Fig. 4. Reporting From System Card

### A. Preparedness Framework

In making risk determinations, OpenAI developed a Preparedness Framework to assess their models. This framework consists of five central elements: tracking catastrophic risk, seeking out unknown unknowns, establishing safety baselines, tasking the preparedness team with work, and creating a cross-functional advisory body. OpenAI also states that its "primary fiduciary duty is to humanity." As a result, they prioritize the alignment of their models. Among the categories of risk they consider, cybersecurity is explicitly included. This categorization is defined in Table I.

There are several concerns regarding this framework and its application. The first is that it is self-policed and self-defined. While it serves as a useful tool for agenda-setting within the community and OpenAI, it remains untested. Additionally, business interests could influence classifications, potentially affecting risk assessments. The framework also lacks external input and consensus, which could result in missed contributions from third parties.

Regarding the internal structure of the framework, it effectively outlines the hierarchy of concerns. The distinctions between low risk and critical risk appear representative of actual threats. However, one key point to consider is the development of high-level exploits. Currently, both research and practical formulations are largely driven by vulnerability discoverability, as seen in SQL injection. However, AI could potentially generate exploits that challenge human capabilities.

| OpenAI Risk Level Definition |
|---|
| **Low** |
| • Only meaningful for non-programming tasks (e.g., translation, summarization). |
| • Vulnerability discovery alone is excluded (harm arises from exploit deployment, not identification). |
| **Medium** |
| • Significantly increases productivity (e.g., >2x time saved) on tasks like converting known exploits into attacks. |
| • Reduces skill barriers, benefiting even lower-level actors using known methods. |
| **High** |
| • Tool-augmented model can develop high-value exploits (e.g., zero-click iOS) without human intervention. |
| • Unblocks many attackers, accelerating professional cyberattacks. |
| **Critical** |
| • Autonomously identifies and develops zero-day exploits across all software projects. |
| • Devises and executes novel end-to-end cyberattacks against hardened targets with only a high-level goal. |

TABLE I
RISK LEVELS AND DEFINITIONS FOR MODEL-ASSISTED CYBERATTACKS

One area of particular concern is binary exploitation. Humans struggle to operate at low levels of abstraction due to the increased cognitive load required for resource management. Meanwhile, AI-powered agents are already being leveraged to discover and exploit vulnerabilities, deploying multi-agent in a coordinated fashion to enhance their effectiveness [13]. As a result, this framework may be misaligned with the wider development community, especially if models are already being used in a manner that would constitute a "high" rating.

## V. APPLICATION OF SDGs

Sustainable Development Goals(SDGs) are less well defined as the framework proposed by OpenAI. However, this provides flexibility as a target. Pursuing SDGs should be taken as a guiding light rather than a concrete plan. This is useful, since the operational context of AI differs significantly. In the domain of cybersecurity, the operational context is largely "cat-and-mouse." With technological developments arising out of an arms raise for each party to out do the other. It is not surprising that more well defined frameworks can become misaligned. Thus, broad strokes are more appropriate for these fields. Several SDG's will applied to derive the benefit of offensive AI as discussed in the aforementioned.

The most appropriate SDGs are 9 (Industry, Innovation and Infrastructure), SDG 16(Peace, Justice, and Strong Institutions), 17( Partnership for the Goals). Through applying these SDGs, the benefit of developing offensive AI can be derived.

### A. SDG 9 (Industry, Innovation and Infrastructure)

The objective of SDG 9 is to promote resilient and sustainable infrastructure, innovation, and industrialization. The targets are sub-goals with the most pertinent being the development of quality, reliable, and sustainable architecture. SDG 9 is important to society as it is a factor in growth and robustness of economies. When considering the development of offensive AI, they play a role in being proactive in detecting vulnerabilities. The value of developing lower risk CTF solvers is thus well grounded with this SDG. However, when considering the development of malware, as is the case with

[7], it is less supported. Malware by design is meant to disrupt control and destroy infrastructure. One could argue that it is creating a stress test for Advanced Persistent Threats (APTs) that lurk in networks, but then the research direction should be different. There is a gap in the offensive abilities presented by malware for AI systems and the ability to mitigate. In [7], the authors use a "Virtual Donkey" as a guardrail using the input and output of a model. The weakness of this method is that is not prepared to meet the demands of adaptive attacks or smuggling using different encoding methods. Thus, there should be a greater emphasis placed on developing defensive mechanisms as there are already vectors for manipulating models. Effectively, the community is already aware of the problems with AI robustness, but we have not caught up enough with the rapidly changing advancements in the AI space.

## B. SDG 16 (Peace, Justice, and Strong Institutions)

The objective of SDG 16 is to promote peaceful, and accountable societies. The relevant targets are to reduce to violence, protect children from abuse, ensure public access to information, promote accountability of institutions, and prevent violence and combat terrorism and crime.

In terms of accountability and violence, one of the most important vectors is at the intersection is protecting data and data breaches. Data breaches relate to one's autonomy and maintaining a sense of liberty, and it is liable to compound over time. Many services use knowledge-based authentication to verify users, which becomes problematic once a critical amount of information is obtained. If a malicious actor acquires enough vital information about a target, they can request additional details and build a complete profile of their target. One infamous instance of this issue occurred with the Internal Revenue Service's "Get Transcript" service. Attackers were able to answer the knowledge-based questions and access people's tax transcripts. This allowed them to gain more personal information and request individual tax returns [14]. Such data can further lead to greater harm. For example, organizations may use this data to manipulate individuals in ways that maximize the organization's own utility [15].Data breaches are a prerequisite for more targeted attacks, allowing hackers to build profiles of victims and potentially steal their identities. Privacy harm is clearly inflicted during these breaches, as hackers typically conduct them with the intent to distribute stolen information for further attacks.

By developing offensive agents to fortify systems, it can reinforce the trust placed within institutions. It also makes it more difficult for hackers to implement further attacks if they are unable to build their supply chain of credentials. Developing malware in itself does not overly support system robustness. It makes administrations aware of specific weakness or payloads, but the defenses implemented mostly act as a deterrent. The most typical defense implemented are system prompt instructions. However, prompt injection techniques can misalign safeguards to manipulate models. As previously argued, there is greater need in defensive research before offensives methods should be made a priority.

## C. SDG 17 (Partnerships For the Goals)

The objective of SDG 17 is focused on promoting global collaboration and gathering resources to achieve the other SDG. This goal is meant to support and catalyze the other goals. Cyber security generally relates to this overarching goal through the robustness of infrastructure and the faith held within institutions. If transactions are secure, then they should flow more smoothly. Using agents to achieve to this goal is beneficial to society. The same logic applies to the faith placed in institutions to safeguard user data. With the current circumstances, it is difficult to determine how malware based initiatives are supportive of these goals.

## VI. CONCLUSION

Two general forms of offensive AI are considered in this research. The first are agents that are designed to solve CTFs, and the other is AI-based malware. Through applying SDGs, designing agents to solve CTF challenges can be seen as more grounded and are generally low risk when applying other frameworks. That is not to say it is without risk. Some of the agents being designed are on a trajectory to design exploits that are difficult for humans to interact with and mitigate, as is the case with low-level exploit development. However, it appears to support the SDGs sufficiently to argue that it should be a priority when considering the risks and rewards of further developments and the potential offered.

As for malware development, it appears to be too dangerous to justify. They do not seem to overtly advance SDGs but carry significant risk when considering OpenAI's framework. While some advances in detection exist, the defenses seem to lack the robustness needed. When observing how the payload blends into the context and the patterns in the activations of models, it appears that greater research should be placed in mechanistic interpretability to make models themselves robust to these attacks rather than continually leveraging external frameworks or system prompts.

## VII. REPRODUCTION

Check out the repository on GitHub: https://github.com/rymarinelli/Wormy_AI/tree/main